\documentclass[twoside]{article}

\usepackage[accepted]{aistats2021}
\usepackage[round]{natbib}

\usepackage{todonotes}

\usepackage{caption}
\usepackage{subcaption}

\usepackage{hyperref}
\usepackage{cleveref}

\usepackage[export]{adjustbox}
\usepackage{booktabs}
\usepackage{listings}
\usepackage{xcolor}
\definecolor{codegreen}{rgb}{0,0.6,0}
\definecolor{codegray}{rgb}{0.5,0.5,0.5}
\definecolor{codepurple}{rgb}{0.58,0,0.82}
\definecolor{backcolour}{rgb}{0.95,0.95,0.92}
\lstdefinestyle{mystyle}{
  backgroundcolor=\color{backcolour},
  commentstyle=\color{codegreen},
  keywordstyle=\color{magenta},
  numberstyle=\tiny\color{codegray},
  stringstyle=\color{codepurple},
  basicstyle=\ttfamily\footnotesize,
  breakatwhitespace=false,
  breaklines=true,
  captionpos=b,
  keepspaces=true,
  numbers=left,
  numbersep=5pt,
  showspaces=false,
  showstringspaces=false,
  showtabs=false,
  tabsize=2
}
\crefname{lstlisting}{listing}{listings}
\Crefname{lstlisting}{Listing}{Listings}

\usepackage{amssymb,amsthm}
\theoremstyle{definition}
\newtheorem{definition}{Definition}[section]

\usepackage[bold]{hhtensor}
\usepackage{mathtools}

\newcommand{\simiid}{\overset{\text{iid}}{\sim}}
\DeclarePairedDelimiterX{\infdivx}[2]{(}{)}{%
  #1\;\delimsize\|\;#2%
}

\DeclareMathOperator{\MB}{MB}
\DeclareMathOperator{\Pa}{Pa}
\DeclareMathOperator{\diag}{diag}

\newcommand{\bE}{\mathbb{E}}
\newcommand{\cN}{\mathcal{N}}

\newcommand{\vx}{\vec{x}}
\newcommand{\vy}{\vec{y}}

\begin{document}

%

%

\twocolumn[

  \aistatstitle{Accelerating Metropolis-Hastings with Lightweight Inference Compilation}

  \aistatsauthor{ Feynman Liang \And Nimar Arora \And Nazanin Tehrani \And Yucen Li \And Michael Tingley \And Erik Meijer }
  \runningauthor{Liang, Arora, Tehrani, Li, Tingley, Meijer}

  \aistatsaddress{ UC Berkeley \And Facebook \And Facebook \And Facebook \And Facebook \And Facebook} ]

\begin{abstract}
  In order to construct accurate proposers for Metropolis-Hastings Markov
  Chain Monte Carlo, we integrate ideas from probabilistic graphical models
  and neural networks in an open-source framework we call Lightweight
  Inference Compilation (LIC). LIC implements amortized inference within an
  open-universe declarative probabilistic programming language (PPL).
  Graph neural networks are used to parameterize proposal distributions
  as functions of Markov blankets, which during ``compilation'' are optimized
  to approximate single-site Gibbs sampling distributions. Unlike prior work
  in inference compilation (IC), LIC forgoes importance sampling of linear
  execution traces in favor of operating directly on Bayesian networks.
  Through using a declarative PPL, the Markov blankets of nodes (which may be
  non-static) are queried at inference-time to produce proposers Experimental
  results show LIC can produce proposers which have less parameters, greater
  robustness to nuisance random variables, and improved posterior sampling in
  a Bayesian logistic regression and $n$-schools inference application.
\end{abstract}

\section{Background}

\begin{figure*}
  \centering
  \includegraphics[width=\linewidth]{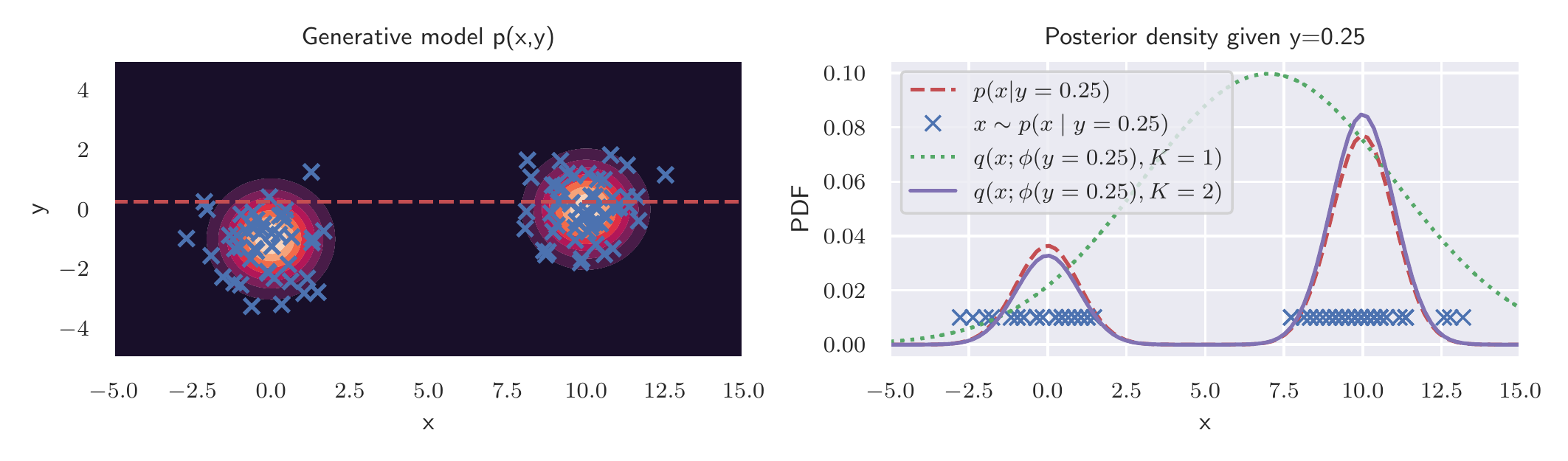}
  \caption{
    Intuition for Lightweight Inference Compilation (LIC).
    LIC uses samples $(x_i, y_i) \overset{\text{iid}}{\sim} p$ (blue ``x'' in left)
    drawn from the joint density $p(x, y)$ to approximate the expected inclusive KL-divergence
    $\bE_{p(y)} D_{\text{KL}}\infdivx{p(x \mid y)}{q(x; \phi(y))}$
    between the posterior $p(x \mid y)$ and the LIC proposal distribution
    $q(x ; \phi(y))$. For an observation $y=0.25$ (dashed red line in left),
    the posterior $p(x \mid y=0.25)$ (dashed red line in right) is ``approximated''
    by samples ``close'' to $y$ (blue ``x'' in right) to form an empirical
    inclusive KL-divergence minimized by LIC. As inclusive KL-divergence
    encourages a mass-covering / mean-seeking fit, the resulting proposal
    distribution $q(x;\phi(y=0.25), K=1)$ (green dotted line in right)
    when using a single ($K=1$) Gaussian proposal density covers both modes
    and can successfully propose moves which cross between the two mixture
    components. Using a 2-component ($K=2$) GMM proposal density results in
    $q(x;\phi(y=0.25), K=2)$ (purple solid line in right) which captures both the
    bi-modality of the posterior as well as the low probability region between
    the two modes.
    As a result of sampling the generative model, LIC can discover
    both posterior modes and their appropriate mixture weights
    (whereas other state of the art MCMC samplers fail, see \cref{fig:gmm_mode_escape}).
  }\label{fig:intuition}
\end{figure*}

Deriving and implementing samplers has traditionally been a high-effort and
application-specific endeavour \citep{porteous2008fast,murray2010elliptical},
motivating the development of general-purpose probabilistic programming
languages (PPLs) where a non-expert can specify a generative model (i.e.
joint distribution) $p(\vx, \vy)$ and the software automatically performs
inference to sample latent variables $\vx$ from the posterior $p(\vx \mid
  \vy)$ conditioned on observations $\vy$. While exceptions exist,
modern general-purpose PPLs typically implement variational inference
\citep{bingham2019pyro}, importance sampling
\citep{wood2014new,le2017inference}, or Monte Carlo Markov Chain (MCMC,
\cite{wingate2011lightweight,tehrani2020beanmachine}).

Our work focuses on MCMC. More specifically, we target lightweight
Metropolis-Hastings (LMH, \cite{wingate2011lightweight}) within a recently developed
declarative PPL called \texttt{beanmachine} \citep{tehrani2020beanmachine}.
The performance of Metropolis-Hastings critically depends on the quality of
the proposal distribution used, which is the primary goal of LIC.
LIC makes the following contributions:
\begin{enumerate}
  \item We present a novel implementation of inference compilation (IC) within an
        open-universe declarative PPL which combines Markov blanket structure with
        graph neural network architectures
  \item We describe how to handle novel data encountered at run-time which
        was not observed during compilation, which is an issue faced by
        any implementation of IC supporting open-universe models.
  \item We demonstrate LIC's ability to escape local modes, improved
        robustness to nuisance random variables, and improvements over
        state-of-the-art methods across a number of metrics in two
        industry-relevant applications
\end{enumerate}

\subsection{Declarative Probabilistic Programming}

To make Bayesian inference accessible to non-exports, PPLs provide
user-friendly primitives in high-level programming languages for abstraction
and composition in model representation
\citep{goodman2013principles,ghahramani2015probabilistic}. Existing PPLs
can be broadly classified based on the representation inference is performed
over, with declarative PPLs \citep{lunn2000winbugs,plummer2003jags,milch20071,tehrani2020beanmachine}
performing inference over Bayesian graphical networks and imperative PPLs
conducting importance sampling \citep{wood2014new} or MCMC
\citep{wingate2011lightweight} on linearized execution traces.
Because an execution trace is a topological sort of an instantiated
Bayesian network, declarative PPLs preserve additional model structure such
as Markov blanket relationships which are lost in imperative PPLs.

\begin{definition}
  The \emph{Markov Blanket} $\MB(x_i)$ of a node $x_i$
  is the minimal set of random variables such that
  \begin{align}
    p(x_i \mid \vx_{-i}, \vy)
     & = p(x_i \mid \MB(x_i))
    \label{eq:gibbs-mb}
  \end{align}
  In a Bayesian network, $\MB(x_i)$ consists of the parents, children, and
  children's parents of $x_i$ \citep{pearl1987evidential}.
\end{definition}


\subsection{Inference Compilation}

Amortized inference \citep{gershman2014amortized} refers to the re-use of
initial up-front learning to improve future queries. In context of
Bayesian inference \citep{marino2018iterative,zhang2018advances} and IC
\citep{paige2016inference,weilbach2019efficient,harvey2019attention},
this means using acceleration performing multiple inferences over
different observations $\vy$ to amortize a one-time
``compilation time.'' While compilation in both trace-based IC
\citep{paige2016inference,le2017inference,harvey2019attention} and LIC
consists of drawing forward samples from the generative model $p(\vx, \vy)$ and
training neural networks to minimize inclusive KL-divergence,
trace-based IC uses the resulting neural network artifacts to parameterize
proposal distributions for importance sampling while LIC uses them for MCMC
proposers.

\subsection{Lightweight Metropolis Hastings}

LMH (\cite{wingate2011lightweight,ritchie2016c3}) updates random variables
within a probabilistic model one at a time according to a Metropolis-Hastings
rule while keeping all the other variables fixed. While a number of choices
for proposal distribution exist, the single-site Gibbs sampler which proposes
from \cref{eq:gibbs-mb} enjoys a 100\% acceptance probability
\citep{pearl1987evidential} and provides a good choice when available
\citep{lunn2000winbugs,plummer2003jags}. Unfortunately, outside of discrete
models they are oftentimes intractable to directly sample so another proposal
distribution must be used. LIC seeks to approximate these single-site Gibbs
distributions using tractable neural network approximations.

\subsection{Related Works}

\begin{figure*}
  \centering
  \includegraphics[width=0.90\linewidth]{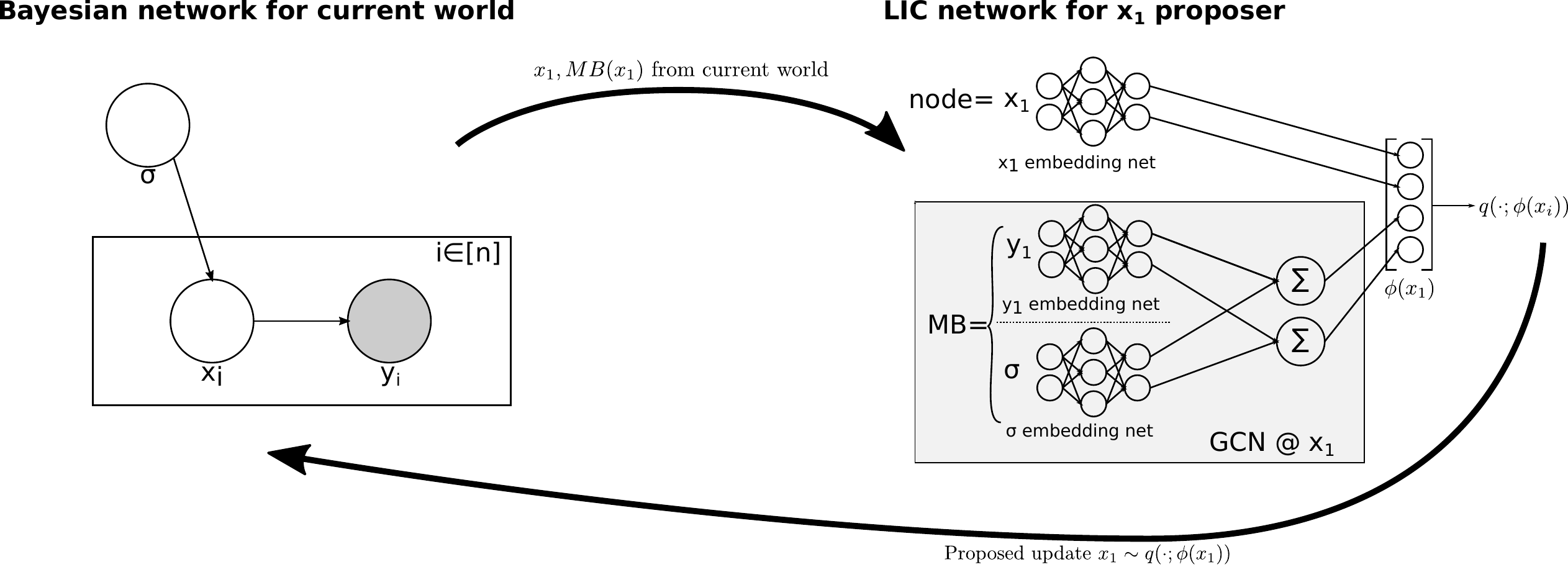}
  \caption{The Markov blankets in the Bayesian network for \cref{eq:conj-normal}
    (left, expressed in plate notation) are available in a declarative PPL,
    and are used as inputs to LIC. The LIC proposer for node $x_1$ (right)
    is obtained by first performing neural network embedding of $x_1$
    and every node in its Markov blanket, followed by a graph convolutional
    network aggregation over the Markov blanket of $x_1$.
    The resulting vectors are then combined to yield a parameter vector $\phi(x_1)$
    for a proposal distribution $q(\cdot; \phi(x_1))$ which is then sampled
    for proposing an update within Metropolis-Hastings.
  }\label{fig:example-ic-arch}
\end{figure*}

Prior work on IC in imperative PPLs can be broadly classified based on the
order in which nodes are sampled. ``Backwards'' methods approximate an inverse
factorization, starting at observations and using IC artifacts to propose
propose parent random variables. Along these lines, \cite{paige2016inference}
use neural autoregressive density estimators but heuristically invert the model
by expanding parent sets. \cite{webb2018faithful} proposes a more principled
approach utilizing minimal I-maps and demonstrate that minimality of inputs to
IC neural networks can be beneficial; an insight also exploited through LIC's
usage of Markov blankets. Unfortunately, model inversion is not possible
in universal PPLs \citep{le2017inference}.

The other group of ``forwards'' methods operate in the same direction as the
probabilistic model's dependency graph. Starting at root nodes, these methods
produce inference compilation artifacts which map an execution trace's prefix
to a proposal distribution. In \cite{ritchie2016deep}, a user-specified
model-specific guide program parameterizes the proposer's architecture and
results in more interpretable IC artifacts at the expense of increased user
effort. \cite{le2017inference} automates this by using a recurrent neural
network (RNN) to summarize the execution prefix while constructing a node's
proposal distribution. This approach suffers from well-documented RNN
limitations learning long distance dependencies
\citep{hochreiter1998vanishing}, requiring mechanisms like attention
\citep{harvey2019attention} to avoid degradation in the presence of long
execution trace prefixes (e.g. when nuisance random variables are present).

With respect to prior work, LIC is most similar to the attention-based
extension \citep{harvey2019attention} of \cite{le2017inference}. Both methods
minimize inclusive KL-divergence empirically approximated by samples from the
generative model $p(\vx, \vy)$, and both methods use neural networks to
produce a parametric proposal distribution from a set of inputs sufficient
for determining a node's posterior distribution. However, important distinctions
include (1) LIC's use of a declarative PPL implies Markov blanket
relationships are directly queryable and ameliorates the need for also
learning an attention mechanism, (2) LIC uses a graph neural network to
summarize the Markov blanket rather than a RNN over the execution trace
prefix, and (3) LIC can handle open-universe models where novel random variables
are encountered at inference time and no embedding network exists.


\section{Lightweight Inference Compilation}

\subsection{Architecture}
\label{ssec:architecture}

\Cref{fig:example-ic-arch} shows a sketch of LIC's architecture.
For every latent node $x_i$, LIC constructs a mapping $(x_i, \MB(x_i)) \mapsto \phi(x_i)$
parameterized by feedforward and graph neural networks to produce a parameter
vector $\phi(x_i)$ for a parametric density estimator $q(\cdot; \phi(x_i))$.
Every node $x_i$ has feedforward ``node embedding network'' used
to map the value of the underlying random variable into a vector space of
common dimensionality. The set of nodes in the Markov blanket are then
summarized to a fixed-length vector following \cref{ssec:dynamic-mb}, and
a feedforward neural network ultimately maps the concatenation of the
node's embedding with its Markov blanket summary to proposal
distribution parameters $\phi(x_i)$.

\subsubsection{Dynamic Markov Blanket embeddings}
\label{ssec:dynamic-mb}


Because a node's Markov blanket may vary in both its size and elements (e.g.
in a GMM, a data point's component membership may change during MCMC),
$\MB(x_i)$ is a non-static set of vectors (albeit all of the same dimension
after node embeddings are applied) and a feed-forward network with fixed
input dimension is unsuitable for computing a fixed-length proposal parameter
vector $\phi(x_i)$. Furthermore, Markov blankets (unlike execution trace
prefixes) are unordered sets and lack a natural ordering hence use of a
RNN as done in \cite{le2017inference} is inappropriate.
Instead, LIC draws motivation from graph neural networks
\citep{scarselli2008graph,dai2016discriminative} which have demonstrated
superior results in representation learning \cite{bruna2013spectral} and
performs summarization of Markov Blankets following \cite{kipf2016semi} by
defining
\begin{align*}
  \phi(x_i) = \sigma\left(
  \matr{W}
  \underset{x_j \in \text{MB}(x_i)}{\square}
  \frac{1}{\sqrt{\lvert \text{MB}(x_i) \rvert \lvert \text{MB}(x_j) \rvert}}
  f_j(x_j)
  \right)
\end{align*}
where $f_j(x_j)$ denotes the output of the node embedding network for node
$x_j$ when provided its current value as an input, $\square$ is any
differentiable permutation-invariant function (summation in LIC's case),
and $\sigma$ is an activation function.

\subsubsection{Parameterized density estimation}

The resulting parameter vectors $\phi(x_i)$ of LIC are ultimately used to
parameterize proposal distributions $q(x_i; \phi(\MB(x_i)))$ for MCMC sampling.
For discrete $x_i$, LIC directly estimates logit scores over the
support. For continuous $x_i$, LIC transforms continuous $x_i$ to
unconstrained space following \cite{carpenter2017stan} and models
the density using a Gaussian mixture model (GMM). Note that although more
sophisticated density estimators such as masked autoregressive flows
\citep{kingma2016improved,papamakarios2017masked} can equally be used.

\subsection{Objective Function}

To ``compile'' LIC, parameters are optimized to minimize the inclusive
KL-divergence between the posterior distributions and inference compilation
proposers: $D_{\text{KL}}\infdivx{p(\vx \mid \vy)}{q(\vx \mid \vy; \phi)}$.
Consistent with \cite{le2017inference}, observations $\vy$
are sampled from the marginal distribution $p(\vy)$, but note that this may
not be representative of observations $\vy$ encountered at inference. The
resulting objective function is given by
\begin{align}
   & \bE_{p(\vy)}\left[
    D_{\text{KL}}\infdivx{p(\vx \mid \vy)}{q(\vx \mid \vy; \phi)}
  \right]                                               \label{eq:e-kl-inc} \\
   & = \bE_{p(\vx, \vy)}\left[
    \log \frac{p(\vx \mid \vy)}{q(\vx \mid \vy; \phi)}
  \right]                                               \nonumber           \\
   & \propto \bE_{p(\vx, \vy)}\left[
    -\log q(\vx \mid \vy; \phi)
  \right]                                               \nonumber           \\
   & \approx \sum_{i=1}^N  -\log q(\vx \mid \vy; \phi),
  \qquad (\vx,\vy) \simiid p(\vx,\vy)                   \nonumber           \\
   & \eqqcolon \mathcal{L}(\phi) \nonumber
\end{align}
where we have neglected a conditional entropy term independent of $\phi$
and performed Monte-Carlo approximation to an expectation.
The intuition for this objective is shown in \Cref{fig:intuition}, which shows
how samples from $p(\vx, \vy)$ (left) form an empirical joint approximation where
``slices'' (at $y=0.25$ in \cref{fig:intuition}) yield posterior approximations
which the objective is computed over (right).

\subsubsection{Open-universe support / novel data at runtime}

Because the number of random variables in an open-universe probabilistic can
be unbounded, it is impossible for any finite-time training procedure which relies
on sampling the generative model $p(\vx, \vy)$ to encounter all possible random variables.
As a consequence, any implementation of IC within a universal PPL must address the issue
of novel data encountered at inference time. This issue is not explicitly addressed
in prior work \citep{le2017inference,harvey2019attention}. In LIC's case,
if a node $x_i$ does not have an existing LIC artifact $x_i \mapsto \phi(x_i)$
then we fall back to proposing from the parent-conditional prior $q(x_i;
  \MB(x_i), \phi_i) = p(x_i \mid \Pa(x_i))$ (i.e. ancestral sampling).

\section{Experiments}

To validate LIC's competitiveness, we conducted experiments benchmarking a variety
of desired behaviors and relevant applications. In particular:
\begin{itemize}
  \item Training on samples from the joint distribution $p(\vx, \vy)$ should enable
        discovery of distant modes, so LIC samplers should be less likely to
        to get ``stuck'' in a local mode. We validate this in \cref{ssec:gmm}
        using a GMM mode escape experiment, where we see LIC escape not only
        escape a local mode but also yield accurate mixture component
        weights.
  \item When there is no approximation error (i.e. the true posterior density is
        within the family of parametric densities representable by LIC), we expect
        LIC to closely approximate the posterior at least for the range of
        observations $\vy$ sampled during compilation (\cref{eq:e-kl-inc})
        with high probability under the prior $p(\vy)$. \Cref{ssec:conj-gaussian}
        shows this is indeed the case in a conjugate Gaussian-Gaussian model
        where a closed form expression for the posterior is available.
  \item Because Markov blankets can be explicitly queried, we expect LIC's
        performance to be unaffected by the presence of nuisance random variables
        (i.e. random variables which extend the execution trace but are statistically
        independent from the observations and queried latent variables). This is
        confirmed in \cref{ssec:nuisance} using the probabilistic program from
        \cite{harvey2019attention}, where we see
        trace-based IC suffering an order of
        magnitude increase in model parameters and compilation time while
        yielding an effective sample size almost $5\times$ smaller
        (\Cref{fig:nuisance}).
  \item To verify LIC yields competitive performance in applications of
        interest, we benchmark LIC against other state-of-the-art MCMC methods
        on a Bayesian logistic regression problem
        (\cref{ssec:blr}) and on a generalization of the classical eight schools
        problem \citep{rubin1981estimation} called $n$-schools
        (\cref{ssec:nschools}) which is used in production at a large internet
        company for Bayesian meta-analysis \cite{sutton2001bayesian}.
        We find that LIC exceeds the performance of adaptive random walk
        Metropolis-Hastings \citep{garthwaite2016adaptive}
        and Newtonian Monte Carlo \citep{arora2020newtonian} and
        yields comparable performance to NUTS \citep{hoffman2014no} despite
        being implemented in an interpreted (Python) versus compiled (C\texttt{++})
        language.
\end{itemize}

\begin{figure*}
  \includegraphics[width=\linewidth]{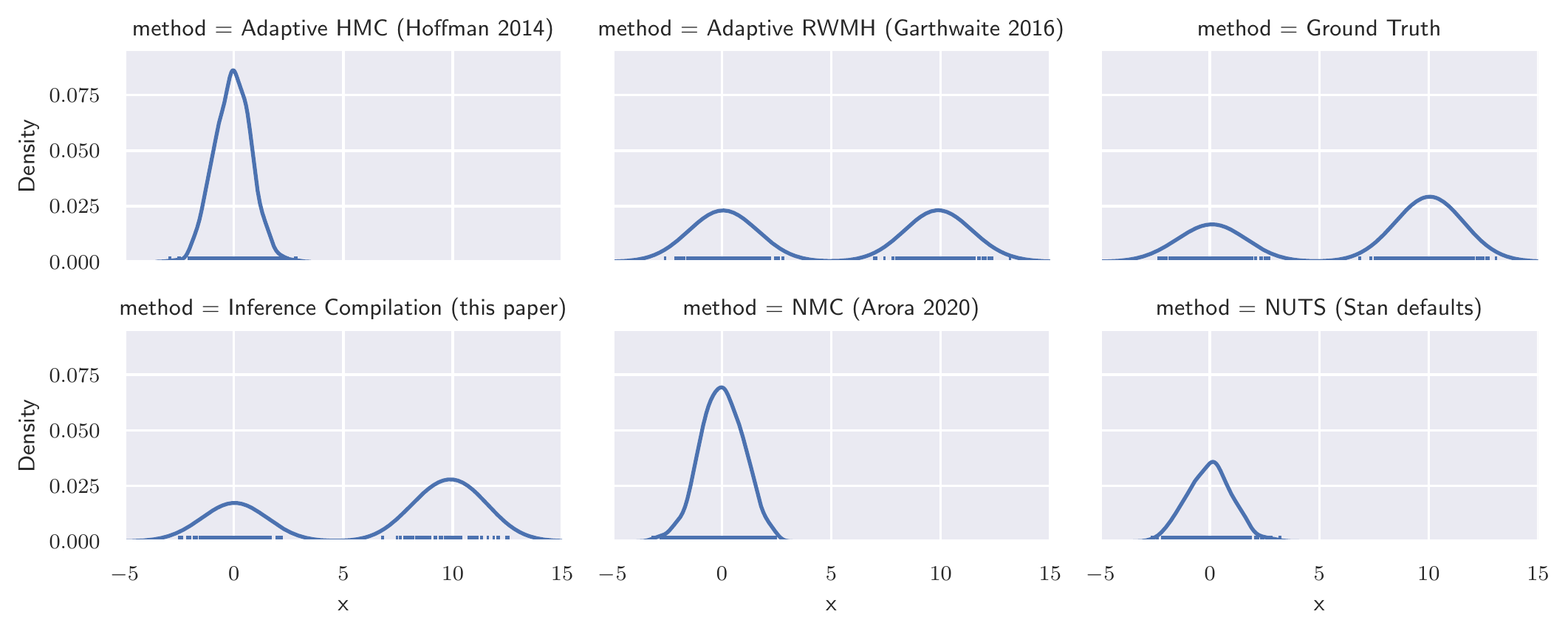}
  \caption{
    When sampling the bi-modal posterior density from \cref{fig:intuition},
    only inference compilation (IC, this paper) and adaptive step-size random walk
    Metropolis-Hastings (Adaptive RWMH, \cite{garthwaite2016adaptive}) are
    able to recover both posterior modes. Whereas RWMH's posterior samples
    erroneously assign approximately equal probability to both modes,
    IC's samples faithfully reproduce the ground truth and yields
    higher probability for the mode at $x=10$ than the mode at $x=0$.
  }\label{fig:gmm_mode_escape}
\end{figure*}

\subsection{GMM mode escape}
\label{ssec:gmm}

Consider the multi-modal posterior resulting from conditioning on $y=0.25$ in
the 2-dimensional GMM in \cref{fig:intuition}, which is comprised of two
Gaussian components with greater mixture probability on the right-hand
component and a large energy barrier separating the two components. Because
LIC is compiled by training on samples from the joint distribution $p(x, y)$,
it is reasonable to expect LIC's proposers to assign high probability to
values for the latent variable $x$ from both modes. In contrast, uncompiled
methods such as random walk Metropolis-Hastings (RWMH) and NUTS may encounter
difficulty crossing the low-probability energy barrier and be unable to escape
the basin of attraction of the mode closest to their initialization.

This intuition is confirmed in \cref{fig:gmm_mode_escape}, which illustrates
kernel density estimates of 1,000 posterior samples obtained by a variety of
MCMC algorithms as well as ground truth samples. HMC with adaptive step size
\citep{hoffman2014no}, NMC, and NUTS with the default settings as implemented
in Stan \citep{carpenter2017stan} are all unable to escape the mode they are
initialized in. While both LIC and RWMH with adaptive step size escape the
local mode, RWMH's samples erroneously imply equal mixture component
probabilities whereas LIC's samples faithfully reproduce a higher component
probability for the right-hand mode.

\subsection{Conjugate Gaussian-Gaussian Model}
\label{ssec:conj-gaussian}

We next consider a Gaussian likelihood with a Gaussian mean prior, a
conjugate model with closed-form posterior given by:
\begin{align}
  x                                 \sim \mathcal{N}(0, \sigma_x),
                                    & \qquad y \sim \mathcal{N}(x, \sigma_y) \label{eq:conj-normal} \\
  \Pr[x \mid y, \sigma_x, \sigma_y] & \sim \mathcal{N}\left(
  \frac{\sigma_y^{-2}}{\sigma_x^{-2} + \sigma_{y}^{-2}} y,
  \frac{1}{\sigma_x^{-2} + \sigma_{y}^{-2}}
  \right) \nonumber
\end{align}
\begin{figure}
  \centering
  \includegraphics[width=0.8\linewidth]{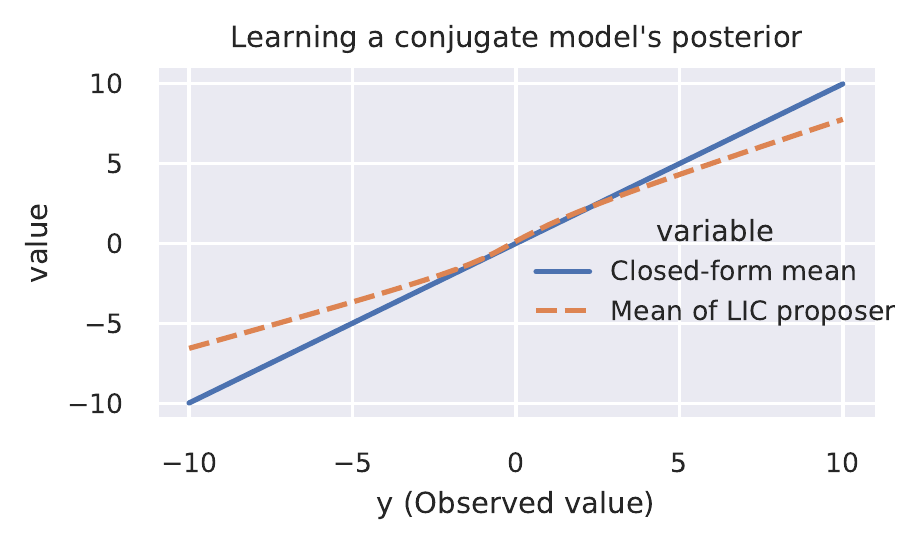}
  \caption{In a conjugate normal-normal model, LIC's proposal distribution
    mean (dashed orange line) closely follows the closed-form posterior mean
    (blue solid line) across a wide range of observed values $y$.}\label{fig:normal-normal}
\end{figure}

There is minimal approximation error because the posterior density is in the
same family as LIC's GMM proposal distributions and the relationship between
the Markov blanket $\MB(x) = \{y\}$ and the posterior mean is a linear
function easily approximated (locally) by neural networks. As a result,
we expect LIC's proposal distribution to provide a good approximation to the
true posterior and LIC to approximately implement a direct posterior sampler.

To confirm this behavior, we trained LIC with a $K=1$ component GMM proposal
density on 1,000 samples and show the resulting LIC proposer's mean as the
observed value $y$ varies in \cref{fig:normal-normal}. Here $\sigma_x = 2$
and $\sigma_y = 0.1$, so the marginal distribution of $y$ (i.e. the
observations sampled during compilation in \cref{eq:e-kl-inc}) is Gaussian
with mean $0$ and standard deviation $\sqrt{\sigma_x^2 + \sigma_y^2} \approx
  2.0025$. Consistent with our expectations, LIC provides a good approximation
to the true posterior for observed values $y$ well-represented during
training (i.e. with high probability under the marginal $p(y)$). While LIC
also provides a reasonable proposer by extrapolation to less represented
observed values $y$, it is clear from \cref{fig:normal-normal} that the
approximation is less accurate. This motivates future work into modifying the
forward sampling distribution used to approximate \cref{eq:e-kl-inc} (e.g.
``inflating'' the prior) as well as adapting LIC towards the distribution of
observations $y$ used at inference time.

\subsection{Robustness to Nuisance Variables}
\label{ssec:nuisance}

An important innovation of LIC is its use of a declarative PPL
and ability to query for Markov blanket relationships so that only statistically
relevant inputs are utilized when constructing proposal distributions.
To validate this yields significant improvement over prior work in IC,
we reproduced an experiment from \cite{harvey2019attention} where
nuisance random variables (i.e. random variables which are statistically independent
from the queried latent variables/observations whose only purpose is to extend the
execution trace) are introduced and the impact on system performance is measured.
As trace-based inference compilation utilizes the execution trace prefix to summarize
program state, extending the trace of the program with nuisance random variables
typically results in degradation of performance due to difficulties encountered
by RNN in capturing long range dependencies as well as the
production of irrelevant neural network embedding artifacts.

We reproduce trace-based IC as described in \cite{le2017inference} using the
author-provided PyProb\footnote{https://github.com/probprog/pyprob} software
package, and implement Program 1 from \cite{harvey2019attention} with the
source code illustrated in \cref{lst:pyprob} where 100 nuisance random
variables are added. Note that although \texttt{nuisance} has no relationship
to the remainder of the program, the line number where they are instantiated
has a dramatic impact on performance. By extending the trace between where
\texttt{x} and \texttt{y} are defined, trace-based IC's RNNs degrade due to
difficulty learning a long-range dependency\citep{hochreiter1998vanishing}
between the two variables. For LIC, the equivalent program expressed in the
\texttt{beanmachine} declarative PPL \citep{tehrani2020beanmachine} is shown
in \cref{lst:bm}. In this case, the order in which random variable
declarations appear is irrelevant as all permutations describe the same
probabilistic graphical model.

\begin{lstlisting}[language=Python,caption={A version of Program 1 from \cite{harvey2019attention} to illustrate nuisance random variables},label={lst:pyprob}]
def magnitude(obs):
  x = sample(Normal(0, 10))
  for _ in range(100):
    nuisance = sample(Normal(0, 10))
  y = sample(Normal(0, 10))
  observe(
    obs**2,
    likelihood=Normal(x**2 + y**2, 0.1))
  return x
\end{lstlisting}

\begin{lstlisting}[language=Python,caption={The equivalent program in beanmachine, where independencies are explicit in program specification},label={lst:bm}]
class NuisanceModel:
  @random_variable
  def x(self):
      return dist.Normal(0, 10)
  @random_variable
  def nuisance(self, i):
      return dist.Normal(0, 10)
  @random_variable
  def y(self):
      return dist.Normal(0, 10)
  @random_variable
  def noisy_sq_length(self):
      return dist.Normal(self.x()**2 + self.y()**2, 0.1)
\end{lstlisting}

\Cref{fig:nuisance} compares the results between LIC and trace-based IC
\citep{le2017inference} for this nuisance variable model. Both \texttt{pyprob}'s
defaults (1 layer 4 dimension sample embedding, 64 dimension address
embedding, 8 dimension distribution type embedding, 10 component GMM
proposer, 1 layer 512 dimension LSTM) and LIC's defaults (used for all
experiments in this paper, 1 layer 4 dimension node embedding, 3 layer 8
dimension Markov blanket embedding, 1 layer node proposal network) with a 10
component GMM proposer are trained on 10,000 samples and subsequently used to
draw 100 posterior samples. Although model size is not directly comparable due
differences in model architecture, \texttt{pyprob}'s resulting models were over $7\times$
larger than those of LIC. Furthermore, despite requiring more than $10\times$ longer
time to train, the resulting sampler produced by \texttt{pyprob} yields
an effective sample size almost $5\times$ smaller than that produced by LIC.
\begin{figure}
  \centering
  \begin{tabular}{lccc}
    \toprule
                           & \# params & compile time & ESS   \\
    \midrule
    LIC (this paper)       & 3,358     & 44 sec.      & 49.75 \\
    \cite{le2017inference} & 21,952    & 472 sec.     & 10.99 \\
    \bottomrule
  \end{tabular}
  \caption{Number of parameters, compilation time (10,000 samples), and effective sample size (100 samples)
    for inference compilation in declarative (beanmachine, this work)
    versus imperative (pyprob, \cite{le2017inference}) PPLs
  }\label{fig:nuisance}
\end{figure}

\subsection{Bayesian Logistic Regression}
\label{ssec:blr}

Consider a Bayesian logistic regression model over $d$ covariates
with prior
$\vec{\beta} \sim \cN_{d+1}(\vec{0}_{d+1}, \diag(10, 2.5\vec{1}_{d}))$
and likelihood $y_i \mid \vx_i \simiid \text{Bernoulli}(\sigma(\vec{\beta}^\top \vx_i))$
where $\sigma(t) = (1 + e^{-t})^{-1}$ is the logistic function.
This model is appropriate in a wide range of classification problems where prior
knowledge about the regression coefficients $\vec{\beta}$ are available.

\begin{figure*}
  \centering
  \caption{
    Results drawing 100 samples across 10 chains (after 1,000 burn-in / adaptation / compilation samples)
    on two Bayesian inference problems.
    The top row of each subfigure aggregates over the 10 chains the (left to
    right) compilation time (time to instantiate the Python class, includes
    neural network training for LIC and Stan's C\texttt{++} codegen and compilation
    for NUTS), inference time (time from invoking \texttt{infer()} to when
    sampling terminates), and the predictive log-likelihood (PLL) on held-out
    test data.
    The bottom row aggregates over both the 10 chains as well as over all
    latent variables the (left to right) estimated expected sample size (ESS,
    higher is better, \cite{geyer2011introduction}) and the rank normalized
    $\widehat{R}$ diagnostic (Rhat, closer to $1$ is better, \cite{vehtari2020rank}).
  }\label{fig:blr_nschools}
  \begin{subfigure}[b]{\textwidth}
    \centering
    \caption{Bayesian logistic regression ($2000$ rows, $10$ features).
      Both LIC and Stan (NUTS) amortize the upfront compilation cost with accelerated inference times.
      LIC achieves comparable PLL to NUTS \cite{hoffman2014no} and other methods,
      and yields ESS comparable to NUTS and higher than any other method.
      The $\widehat{R}$ of LIC is close to $1$ (similar to NUTS and lower than
      all other methods).
    }\label{fig:blr}
    \includegraphics[width=\linewidth]{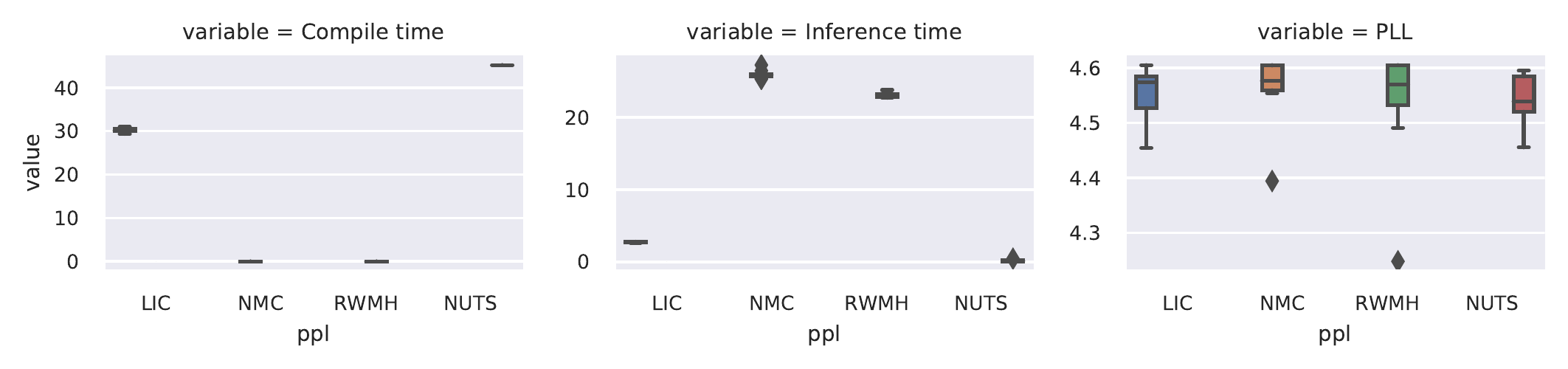}
    \includegraphics[width=0.6\linewidth]{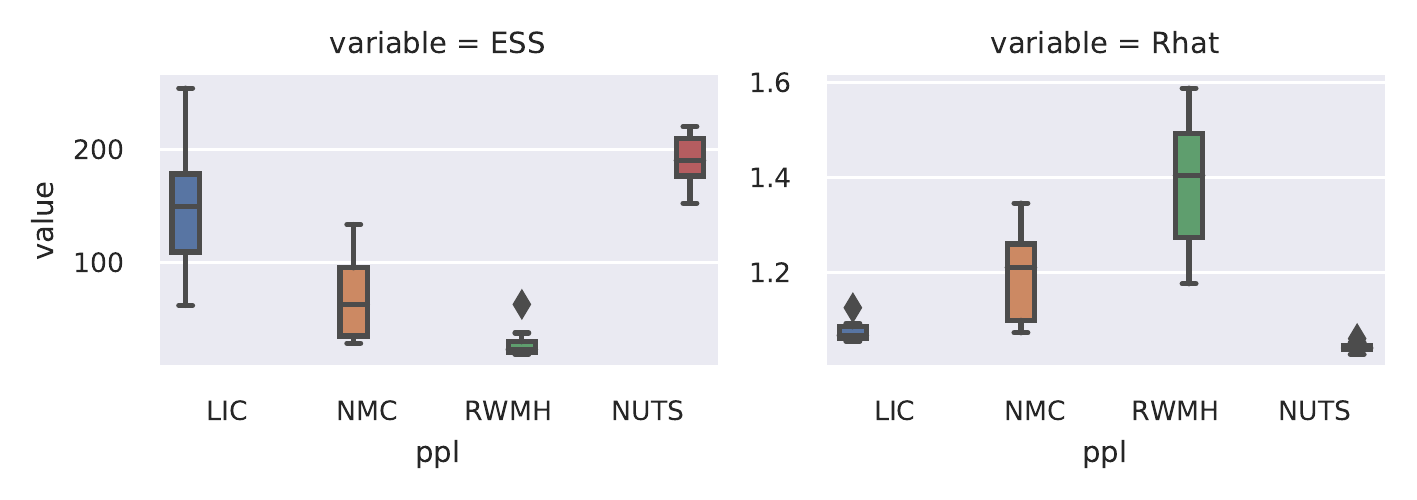}
  \end{subfigure}
  \begin{subfigure}[b]{\textwidth}
    \centering
    \caption{n-schools (1000 schools, 8 states, 5 districts, 5 types).
      Again, increased compilation times are offset by accelerated inference times.
      In this case, LIC achieves comparable PLL to NUTS while simultaneously producing
      higher ESS and lower $\widehat{R}$, suggesting that the resulting posterior samples
      are less autocorrelated and provide a more accurate posterior approximation.
    }\label{fig:nschools}
    \includegraphics[width=\linewidth]{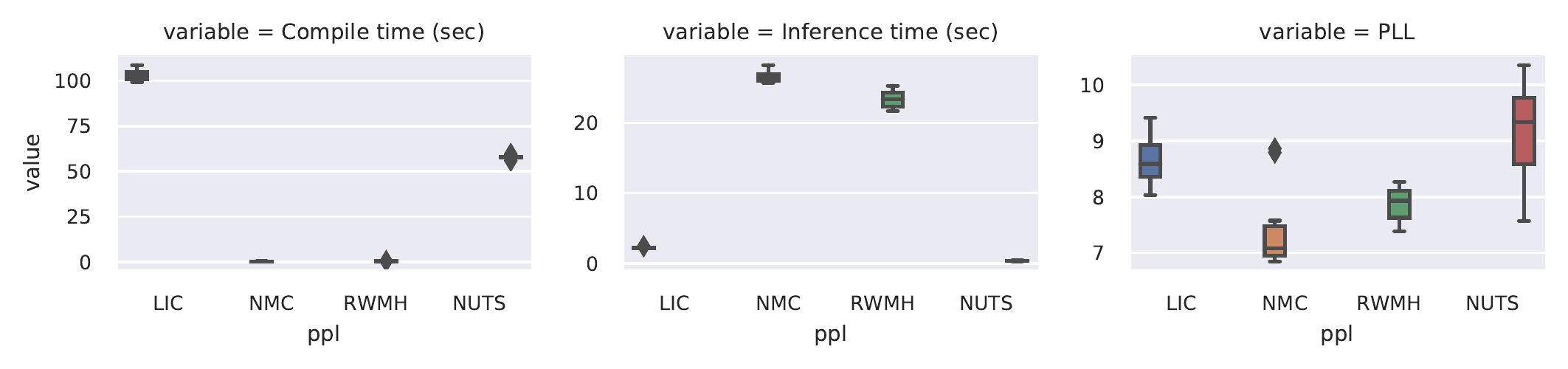}
    \includegraphics[width=0.66\linewidth]{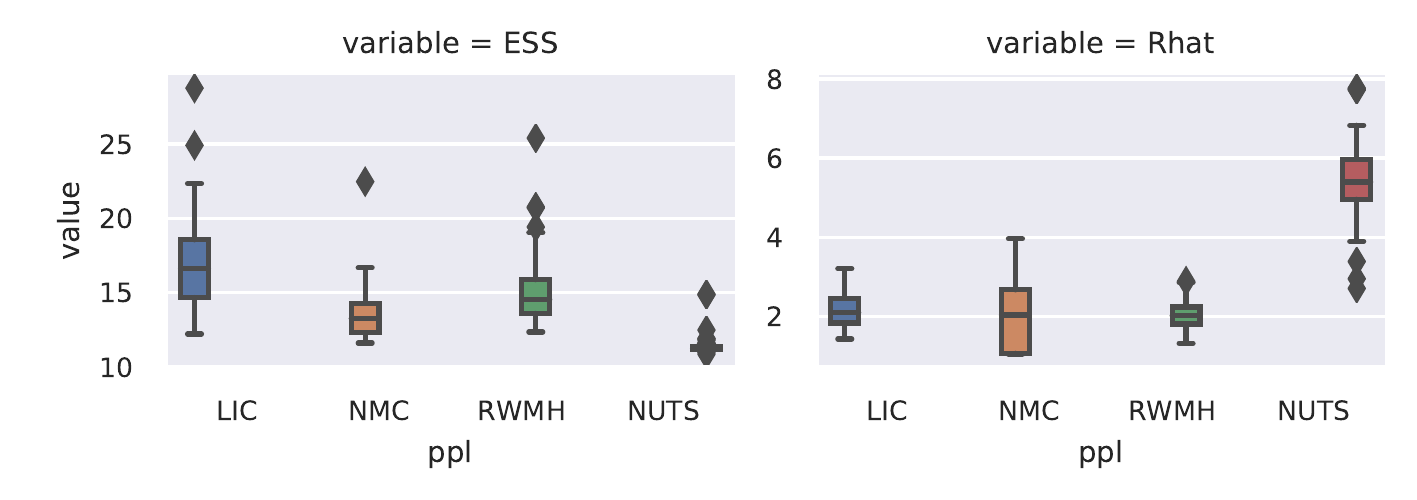}
  \end{subfigure}
\end{figure*}

\Cref{fig:blr} shows the results of performing inference using LIC compared
against other MCMC inference methods. All methods yield similar
predictive log-likelihoods on held-out test data, but LIC and NUTS yield
significantly higher ESS and $\widehat{R}$s closer to $1.0$ suggesting better
mixing and more effective sampling.

\subsection{n-Schools}
\label{ssec:nschools}

The eight schools model \citep{rubin1981estimation} is a Bayesian
hierarchical model originally used to model the effectiveness of schools at
improving SAT scores. $n$-schools is a generalization of this model from $8$ to
$n$ possible treatments, and is used at a large internet company
for performing Bayesian meta-analysis \cite{sutton2001bayesian} to estimate effect sizes
across a range of industry-specific applications.
Let $K$ denote the total number of schools, $n_j$ the number of
districts/states/types, and $j_k$ the district/state/type of school $k$.
\begin{align*}
  \beta_0     & \sim \text{StudentT}(3, 0, 10)                                                                         \\
  \tau_{i}    & \sim \text{HalfCauchy}(\sigma_i) \qquad \text{for}~i \in [\text{district}, \text{state}, \text{type}]  \\
  \beta_{i,j} & \sim \cN(0, \tau_i)  \qquad \text{for}~i \in [\text{district}, \text{state}, \text{type}], j \in [n_i] \\
  y_k         & \sim \cN(\beta_0 + \sum_i \beta_{i,j_k}, \sigma_k)
\end{align*}
\Cref{fig:nschools} presents results in a format analogous to
\cref{ssec:blr}. Here, we see that while both LIC and NUTS yield higher PLLs
(with NUTS outperforming LIC in this case), LICs ESS is significantly higher
than other compared methods. Additionally, the $\widehat{R}$ of NUTS is also
larger than the other methods which suggests that even after 1,000 burn-in samples
NUTS has still not properly mixed.

\section{Conclusion}

We introduced Lightweight Inference Compilation (LIC) for building high
quality single-site proposal distributions to accelerate Metropolis-Hastings
MCMC. LIC utilizes declarative probabilistic programming to retain graphical
model structure and graph neural networks to approximate single-site Gibbs
sampling distributions. To our knowledge, LIC is the first proposed method
for inference compilation within an open-universe declarative probabilistic
programming language and an open-source implementation will be released in
early 2021. Compared to prior work, LIC's use of Markov blankets resolves the
need for attention to handle nuisance random variances and yields posterior
sampling comparable to state-of-the-art MCMC samplers such as NUTS and
adaptive RWMH.


\bibliography{refs.bib}

\end{document}